\documentclass[10pt,twocolumn,letterpaper]{article}

\usepackage{cvpr}
\usepackage{times}
\usepackage{epsfig}
\usepackage{graphicx}
\usepackage{amsmath}
\usepackage{amssymb}
\usepackage{multirow}
\usepackage{bm}
\usepackage{color}
\usepackage{siunitx}
\usepackage{comment}

\usepackage[breaklinks=true,bookmarks=false]{hyperref}

\cvprfinalcopy 

\def\cvprPaperID{2205} 

\begin{document}

\title{Student Becoming the Master:
Knowledge Amalgamation for Joint Scene Parsing, Depth Estimation, and More}


\author{Jingwen Ye\textsuperscript{1}, Yixin Ji\textsuperscript{1}, Xinchao Wang\textsuperscript{2}, Kairi Ou\textsuperscript{3}, Dapeng Tao\textsuperscript{4}, Mingli Song\textsuperscript{1}\\
\textsuperscript{1}College of Computer Science and Technology, Zhejiang University, Hangzhou, China\\
\textsuperscript{2}Department of Computer Science, Stevens Institute of Technology, New Jersey, United States \\
\textsuperscript{3}Alibaba Group, Hangzhou, China\\
\textsuperscript{4}School of Information Science and Engineering, Yunnan University, Kunming, China\\
\{yejingwen, jiyixin, brooksong\}@zju.edu.cn, xinchao.w@gmail.com, \\
suzhe.okr@taobao.com, dptao@ynu.edu.cn}
\maketitle
\pagestyle{empty}
\thispagestyle{empty}
\begin{abstract}
In this paper, we investigate a novel deep-model reusing task. Our goal is to train a lightweight and versatile student model, without  human-labelled annotations, that amalgamates the knowledge and masters the expertise of two pre-trained teacher models working on heterogeneous problems, one on scene parsing and the other on depth estimation. To this end, we propose an innovative training strategy that learns the parameters of the student intertwined with the teachers, achieved by ``projecting''  its amalgamated features onto each teacher's domain and computing the loss.  We also introduce two options to generalize the proposed training strategy to handle three or more tasks simultaneously. The proposed scheme yields very encouraging results. As demonstrated on several benchmarks, the trained student model achieves results even superior to those of the teachers in their own expertise domains and on par with the state-of-the-art fully supervised models relying on human-labelled annotations.
\end{abstract}

\section{Introduction}
Deep learning has accomplished unprecedentedly encouraging results in almost every dimension of computer vision applications. The state-of-the-art performances, however, often come at the price of the huge amount of annotated data trained on clusters for days or even weeks.
In many cases, it is extremely cumbersome to conduct the training on personal workstations with one or two GPU cards, not to mention the infeasibility in the absence of training annotations.

This dilemma has been in part alleviated by the fact that many trained deep models have been released online by developers,  enabling their direct deployment by the community. As such, a series of  work has been conducted to investigate reusing  pre-trained deep models.
Examples include the seminal work of knowledge distilling~\cite{hinton2015distilling}, which learns a compact student model using the soft targets obtained from the teachers. The more recent work of~\cite{yim2017gift} makes one step further by training a student with faster optimization and enhanced performance. Despite the very promising results achieved, existing knowledge distilling methods focus on training a student to handle the same task as a teacher does or a group of homogeneous teachers do.

We propose in this paper to study a related yet new and more challenging model reusing task. Unlike the conventional setup of knowledge distilling where one teacher or an ensemble of teachers working in the same domain, like classification, are provided as input, the proposed task assumes that we are given a number of task-wise heterogeneous teachers, each of which works on a different problem. Our goal is then to train a versatile student model, with a size smaller than the ensemble of the teachers, that \emph{amalgamates} the knowledge and learns the expertise of all teachers, again without accessing  human-labelled annotations.

Towards this end, we look into two vital pixel-prediction applications, scene parsing and depth estimation. We attempt to learn a light student model that simultaneously handle both tasks from two pre-trained teachers, each of which specializes in one task only.
Such a compact and dual-task student model finds its crucial importance in autonomous driving and robotics, where the model to be deployed should, on the one hand, be sufficiently lightweight to run on the edge side, and on the other hand, produce accurate segmentation and depth estimation for self-navigation.

To amalgamate knowledge from the two teachers in different domains, we introduce an innovative block-wise training strategy. By feeding unlabelled images to the teachers and the student, we learn the amalgamated features within each block of the student  intertwined with the teachers. Specifically, we ``project'' the amalgamated features onto each of the teachers' domains to derive the transferred features, which then replace the features in the corresponding block of the teacher network for computing the loss.
As the first attempt along this line, we assume for now that teacher models share the same architecture. This may sound a strong assumption but in fact not, as the encoder-decoder-like architecture has been showing state-of-the-art performance in many vision applications.

We also show that the proposed amalgamation method can be readily generalized for training students with three or more heterogeneous-task teachers. We introduce two such multi-teacher amalgamation strategies, one offline and one online, and demonstrate them with a third task of surface normal estimation.

The proposed training strategy for knowledge amalgamation yields truly promising results. As demonstrated on several benchmarks, the student model ends up with not only a compact size, but also desirable performances superior to those of the teachers on their own domains, indicating the knowledge aggregated from the heterogeneous domains may benefit each other's task.
Without accessing human-labelled annotations, the student model achieves results on par with the fully supervised state-of-the-art models trained with  labelled annotations.

Our contribution is therefore an innovative knowledge amalgamation strategy for training a compact yet versatile student using heterogeneous-task teachers specializing in different domains. We start with the tasks of scene parsing and depth estimation, and show that the proposed strategy can be seamlessly extended to multiple tasks.  Results on several benchmarks demonstrate  the learned student is competent to handle all the tasks of the teachers with superior and state-of-the-art results, but comes with a lightweight size.

\section{Related Work}
We give a brief review here of the recent advances in scene parsing and depth estimation. We also discuss a related but different task,  knowledge distillation, that aims to train a student model handling the same task as the teacher does.

\textbf{Scene Parsing.}
Convolutional neural networks (CNNs) have recently achieved state-of-the-art scene parsing performances and have become the mainstream model. Many variants have been proposed based on CNNs. For example, PSPNet~\cite{zhao2017pyramid} takes advantage of pyramid pooling operation to acquire multi-scale features,  RefineNet~\cite{lin2017refinenet:} utilizes a multi-path structure to exploit features at multiple levels, and FinerNet~\cite{ye2018finer} cascades a series of networks to produce parsing maps with different granularities.  SegNet~\cite{badrinarayanan2017segnet}, on the other hand, employs an encoder-decoder architecture followed by a final pixelwise classification layer. Other models like the mask-based networks of~\cite{he2017mask, pinheiro2015learning, pinheiro2016learning} and the GAN-based ones of~\cite{ kozinski2017an, luc2016semantic, souly2017semi} have also produced very promising scene parsing results.

In our implementation, we have chosen SegNet as our scene-parsing teacher model due to its robust and state-of-the-art performance. However, the  training strategy proposed in Sec.~\ref{sec:method} is  not limited to SegNet only, and other encoder-decoder scene parsers can be adopted as well.


\textbf{Depth Estimation.}
Earlier depth estimation methods~\cite{saxena2005learning,Saxena2007Learning,scharstein2003high-accuracy} rely on handcrafted features and graphical models. For example,  the approach of~\cite{Saxena2007Learning} focuses on outdoor scenes and formulates the depth estimation as a Markov Random Field (MRF) labeling problem, where features are handcrafted. More recent approaches~\cite{laina2016deeper, li2018monocular, liu2015deep} apply CNNs to learn discriminant features automatically, having accomplished very encouraging results. For example, ~\cite{eigen2014depth} introduces a multi-scale deep network, which first predicts a coarse global output followed by finer ones.

{The approaches of~\cite{guillemaut2012space-time, park2017joint} handle  depth estimation   with other vision tasks like segmentation and surface normal prediction.} The work of~\cite{xu2018pad-net:}, on the other hand, proposes a multi-task prediction and distillation network (PAD-Net) structure for joint depth estimation and scene parsing in aim to improve both tasks. Unlike existing approaches, the one proposed in this paper aims to train a student model by learning from two pre-trained teachers working in different domains, without manually-labelled annotations.



\textbf{Knowledge Distillation.}
Existing knowledge distillation methods focus on learning a student model from a teacher or a set of homogeneous teachers working on the same problem. The learned student model is expected to handle the same task as the teacher does, yet comes in a smaller size and preserves the teachers' performance. The work of~\cite{hinton2015distilling}  shows knowledge distilling yields promising results with a regularized teacher model or an ensemble of teachers, when applied to classification tasks. ~\cite{romero2015fitnets:} extends this idea to enable the training of a student that is deeper and thinner than the teacher by using both the outputs and the intermediate representations of the teacher.

Similar to knowledge distillation,~\cite{gao2017knowledge} proposes a multi-teacher single-student knowledge concentration method to classify 100K object categories.
The work of~\cite{shen2019amalgamating} proposes to train a student classifier that handles the comprehensive classification problem by learning from multiple teachers working on different classes.



\begin{figure*}[t]
\centering
\includegraphics[scale = 0.68]{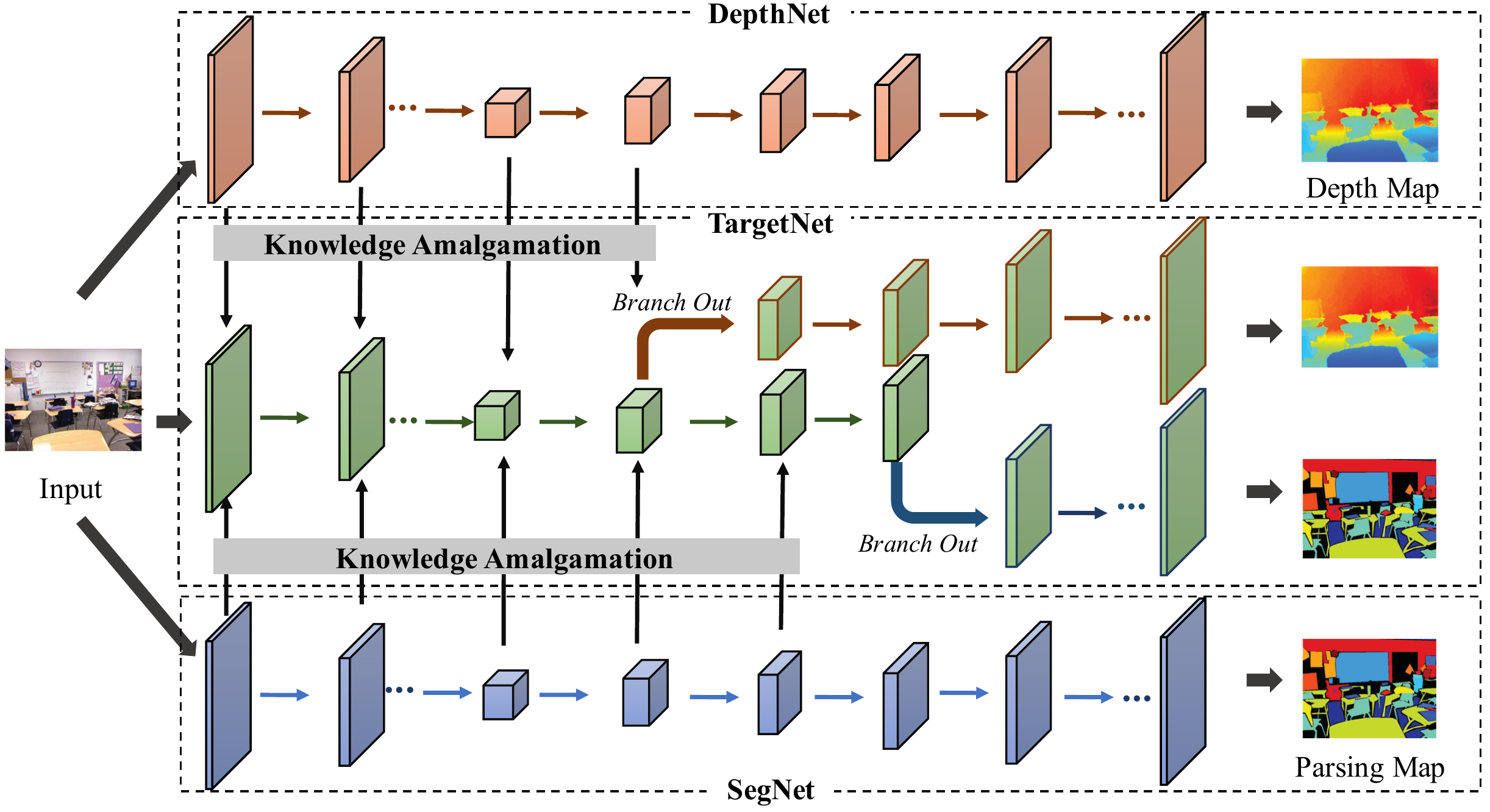}
\caption{ The proposed knowledge amalgamation approach for semantic segmentation and depth estimation with the encoder-decoder network. The \emph{branch out} takes place in the decoder part of the target network.}
\vspace{-3mm}
\label{fig:multi}
\end{figure*}

Apart from classification, knowledge distillation has been utilized in  other tasks~\cite{chen2017learning,huang2018knowledge, xu2018pad-net:}. The work of~\cite{chen2017learning} resolves knowledge distilling on object detection and learns a better student model. The approach of~\cite{huang2018knowledge} focuses on sequence-level knowledge distillation and has  achieved encouraging results  on speech applications.

Unlike  knowledge distilling methods aiming to train a student model that works on the same problem as the teacher does, the proposed knowledge amalgamation methods learn a student model that acquires the super knowledge
of all the heterogeneous-task teachers, each of which specializes in a different domain. Once trained, the student is therefore competent to simultaneously handle various tasks covering the expertise of all teachers.


\section{Problem Definition}
The problem we address here is to learn a compact student model, which we call TargetNet, without human-labeled annotations, that amalgamates knowledge and thus simultaneously handles several different tasks by learning from two or more teachers, each of which specializes in only one task.
Specifically, we focus on two important pixel-prediction tasks, depth estimation and scene parsing, and describe the proposed strategy for knowledge amalgamation in Sec.~\ref{sec:method}. We also show that the proposed training strategy can be readily extended to train a student that handles three or even more tasks jointly.



For the rest of this paper, we take a \emph{block} to be the part of a network bounded by two pooling layers, meaning that in each network all the features maps within a block have the same resolution. We use $I$ to denote an input image, and use ${F_\text{s}^n}$ and ${F_\text{d}^n}$ to respectively denote the feature maps in the $n$-th block of the pre-trained segmentation teacher network~(SegNet) and the depth prediction teacher network~(DepthNet); we also let $S$  denote the final prediction of SegNet and $D$  denote that of DepthNet, and let  $S_i$ and $D_i$  denote the respective predictions at pixel $i$.
In demonstrating the feasibility of  amalgamation from more than two teachers, we look at a third pixel-wise prediction task, surface normal estimation, for which a surface-normal teacher network~(NormNet) is pre-trained. We use $M$ to denote its  predicted normal map with $M_i$ being its $i$-th pixel estimation.

\section{ Pre-Trained Teacher Networks}
\label{sec:teachers}
We describe here the pre-trained teacher networks, SegNet, DepthNet, and NormNet, based on which we train our student model, TargetNet.

Admittedly, we assume  for now that the teacher networks share the same encoder-decoder architecture, but not restricted to any specific design. This assumption might sound somehow strong but in fact not, as many state-of-the-art pixel-prediction models deploy the encoder-decoder architecture. 
Knowledge amalgamation from multiple teachers with arbitrary architectures is left for future work.



\textbf{Scene Parsing Teacher (SegNet).}
The goal of scene parsing is to assign a label denoting a category to each pixel of an image. In our case, we adopt the state-of-the-art SegNet~\cite{badrinarayanan2017segnet}, which has a encoder-decoder architecture, as the scene parsing teacher.
The pixel-wise loss function is taken to be
\begin{equation}
\mathcal{L}_\text{seg}(S^\text{gt},S^*)=\frac{1}{N} \sum_{i}\ell(S_{i}^*,S_i^\text{gt})+\lambda R(\theta),
\label{eq:segloss}
\end{equation}
where
$S_{i}^*$ is the prediction for pixel $i$, $S_{i}^{gt}$ is the ground truth
, $\ell(\cdot)$ is the cross-entropy loss, $N$ is the total pixel number of the input image, and $R$ is the L2-norm regularization term.

Since we do not have access to the human-labeled annotations, we take the prediction $S_i$, converted from $S_i^*$ by one-hot coding, as the supervision for training the TargetNet.

\textbf{Depth Estimation Teacher (DepthNet).}
As another well-studied pixel-level task, depth estimation aims to predict for each pixel a value denoting the depth of an object with respective to the camera. The major difference between scene parsing and depth estimation is, therefore, the  output of the former task is a discrete label while that of the latter is a continuous and positive number.


We transfer depth estimation to a classification problem, which has been proved to be effective~\cite{mousavian2016joint}, by quantizing the depth values to $N_d$ bins, each of which has a length of $l$. For each bin $b$, the network predicts $p(b|x(i))=\exp^{r_i}/\sum_b\exp^{r_{i,b}}$, the probability of having an object at the center of that bin, where $r_{i,b}$ is the response of network at pixel $i$ and bin $b$. The continuous depth value $D_i$ is then computed as:

\begin{equation}
D_i=\sum_{b=1}^{N_d}b\times l\times p(b|x(i)).
\end{equation}
The loss function for depth estimation is computed taken to be:
\begin{equation}
\label{eq:depthloss}
\begin{split}
\mathcal{L}_\text{depth}(D^\text{gt},D)=\frac{1}{N}\sum_{i}(d_i)^2-\frac{1}{2N^2}(\sum_{i}d_i)^2,
\end{split}
\end{equation}
where $d=D^\text{gt}-D$, and $N$ is the total number of valid pixels (we mask out pixels where the ground truth is missing). The prediction of the DepthNet,  $D$, is utilized as the supervision to train the TargetNet.

\textbf{Surface-Normal Prediction Teacher (NormNet).}
Given an input image, the  goal of surface normal prediction is to estimate a surface-normal vector $(x,y,z)$ for each pixel. The loss function to train a surface-normal-prediction teacher network~(NormNet) is taken to be\begin{equation}
\mathcal{L}_\text{norm}(M^\text{gt},M)= -\frac{1}{N}\sum_iM_i\cdot M_i^\text{gt}=-\frac{1}{N}M\cdot M^\text{gt},
\end{equation}
where $M$, $M^\text{gt}$ are respectively the prediction of NormNet and the ground truth, and $M_i$, $M^\text{gt}_i$ are those at pixel $i$.

\section{Proposed Method}
\label{sec:method}
\begin{figure*}[t]
\centering
\includegraphics[scale = 0.67]{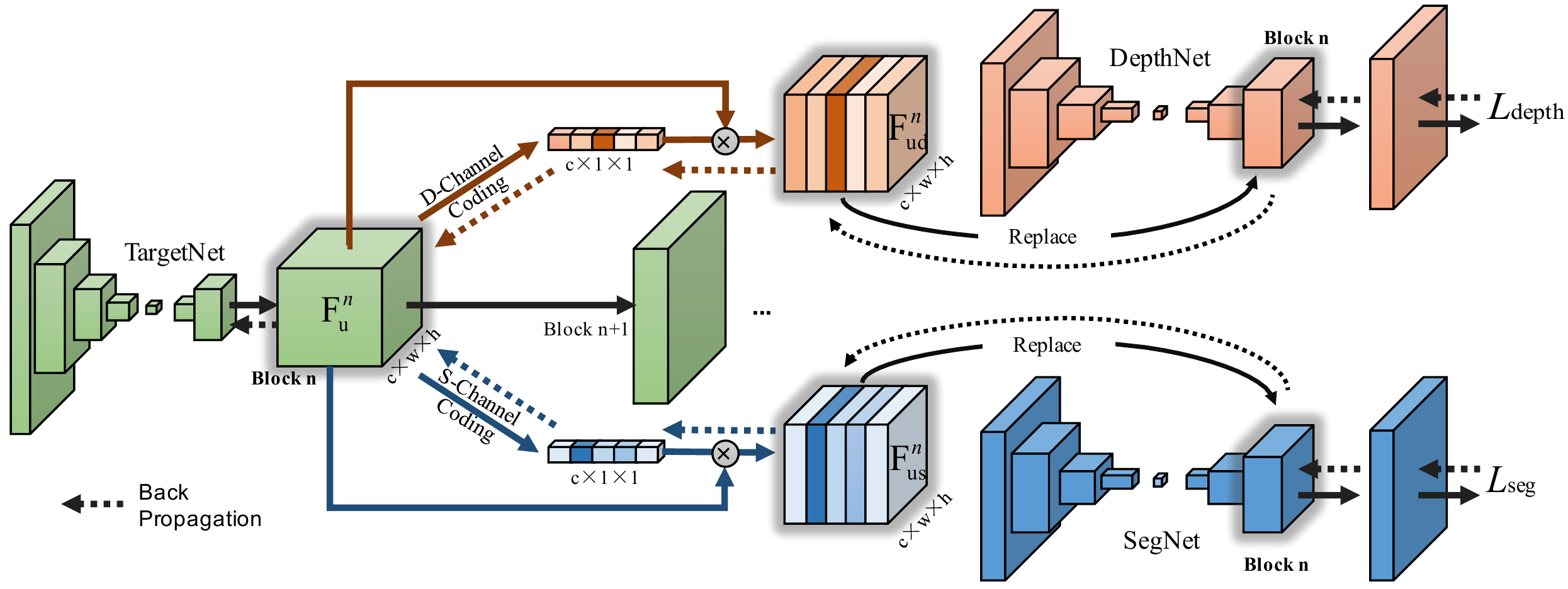}
\caption{ The proposed knowledge amalgamation module within block $n$ of TargetNet. This operation is repeated for every block.}
\vspace{-3mm}
\label{fig:amalgamation}
\end{figure*}

In this section, we describe the proposed approach to learning a compact student network, TargetNet.
%
We introduce a novel strategy that trains the student intertwined with the teachers. At the heart of our approach is a block-wise learning method shown in Fig.~\ref{fig:multi} that learns the parameters of the student network, by ``projecting'' the amalgamated knowledge of the student to each teacher's expertise domain for computing the loss and updating the parameters, as depicted in Fig.~\ref{fig:amalgamation}

In what follows, we start from the amalgamation of the SegNet and DepthNet, and then extend to the amalgamation of multiple networks including surface normal prediction networks~(NormNet).

\subsection{Learning from Two Teachers}
Given the two pre-trained teacher networks, SegNet and DepthNet described in Sec.~\ref{sec:teachers}, we train a compact student model of a similar encoder-decoder architecture, except that the decoder part \emph{eventually} comprises two streams, one for each task as shown in Fig.~\ref{fig:multi}. In principle, all standard backbone CNNs like~AlexNet~\cite{krizhevsky2012imagenet}, VGG~\cite{Simonyan2015Very} and ResNet~\cite{he2016identity} can be employed for constructing such encoder-decoder architecture. In our implementation, we choose the one of  VGG, as done for~\cite{badrinarayanan2017segnet}.


We aim to learn a student model small enough in size so that it is deployable on edge systems, but not smaller as it is expected to master the expertise of both teachers. To this end, for each block of the TargetNet, we take its feature maps to be of the same size as those in the corresponding block of the teachers.

The encoder and the decoder of TargetNet play different roles in the tasks of joint parsing and depth estimation. The encoder part works as a feature extractor to derive
discriminant features for both tasks.
The decoder, on the hand other, is expected to ``project'' or transfer the learned features into each task domain so that they can be activated in different ways for the specific task flow. 

Despite the TargetNet eventually has two output streams, it is initialized with the same encoder-decoder architecture as those of the teachers. We then train the TargetNet and finally branch out the two streams for the two tasks, after which the initial decoder blocks following the branch-out are removed. The overall training process is summarized as follows:
\begin{itemize}
\item {\bf Step~1:} Initialize TargetNet with the same architecture as those of the teachers, as described in~\ref{sec:studentStruct}.
\item {\bf Step~2:} Train each block of TargetNet intertwined with the teachers, shown in Fig.~\ref{fig:amalgamation} and described in Sec.~\ref{sec:blockTrain}.
\item {\bf Step~3:} Decide where to {branch out} on TargetNet, as described in Sec.~\ref{sec:branchout}.
\item {\bf Step~4:} Take the corresponding blocks from the teachers as the branched-out blocks of the student; remove all the initial blocks following the block where the last branch out takes place; fine-tune TargetNet.
\end{itemize}

In what follows, we describe the architecture of the student network, its loss function and training strategy, as well as the branch out strategy.


\subsubsection{TargetNet Architecture}
\label{sec:studentStruct}
The TargetNet is initialized with the same encoder-decoder architecture as those of the teachers, where the structures of the encoder and the decoder are symmetric, as shown in Fig.~\ref{fig:multi}. Each block in the encoder comprises 2~or~3 convolutional layers with $3\times 3$ kernel size, followed with a $2\times 2$ non-overlapping max pooling layer, while in the decoder a pooling layer is replaced by an up-sampling one.

We conduct knowledge amalgamation for each block of TargetNet, by learning the parameters intertwined with the teachers. Let $F_\text{u}^n$ denote the amalgamated feature maps at block~$n$ of TargetNet. We expect  $F_\text{u}^n$ to encode both the parsing and depth information, obtained from the two teachers. To allow the $F_\text{u}^n$ to interact with the teachers and to be updated, we introduce two channel-coding branches, termed \emph{D-Channel Coding} and \emph{S-Channel Coding}, respectively for depth estimation and scene parsing as depicted in Fig.~\ref{fig:amalgamation}. Intuitively,  $F_\text{u}^n$ can be thought as a container of the whole set of features, and can be projected or transformed into the two task domains via the two channels. Here, we use  $F_\text{ud}$ and $F_\text{us}$ to respectively denote the obtained features after passing $F_\text{u}^n$ through the D-Channel and S-Channel Coding, in other words, the projected features in the two task domains.

\begin{figure}[t]
\centering
\includegraphics[scale = 0.5]{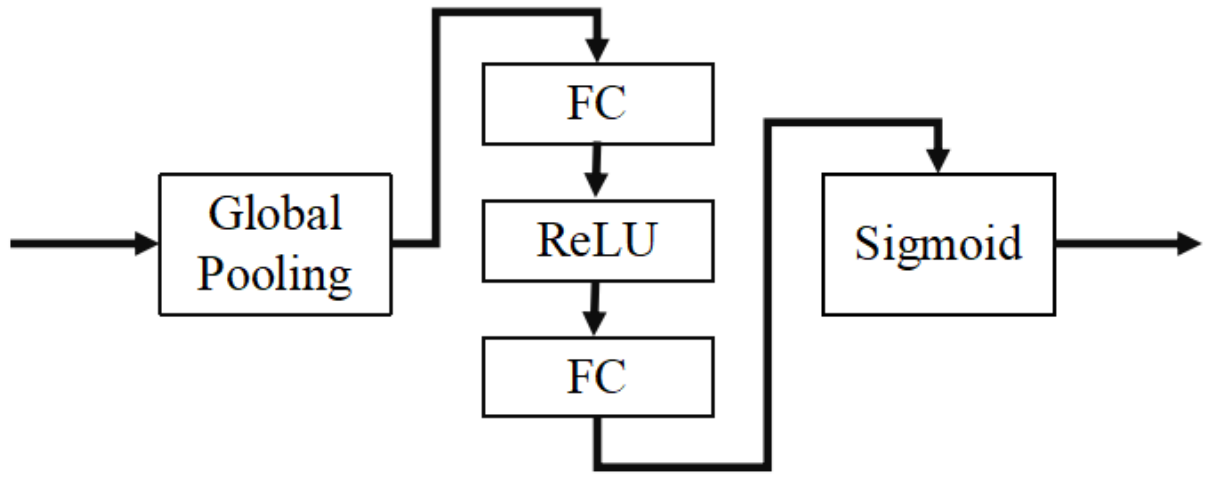}
\caption{ The architecture of the S-Channel Coding and D-Channel Coding.}
\label{fig:channelcoding}
\vspace{-1em}
\end{figure}

The architecture of the two channel coding is depicted in~Fig.~\ref{fig:channelcoding}. Modified from the channel attention module by~\cite{hu2018squeeze-and-excitation}, it consists of a global pooling layer and two fully connected layers, and is very light in size. In fact, it increases the total number of parameters by less than {$4\%$}
only, leading to very low computation cost.

\subsubsection{Loss Function and Training Strategy}
\label{sec:blockTrain}
To learn the amalgamated features $F_\text{u}^n$ at the $n$-th block of TargetNet, we feed the unlabelled samples into both teachers and TargetNet, in which way we obtain the features of two teachers and the initial ones of the Target. Let $F_\text{d}^n$, $F_\text{s}^n$ denote the obtained features at the $n$-th block of the teacher networks, DepthNet and SegNet. One option to learn $F_\text{u}^n$ is to minimize the loss between its projected features $F_\text{ud}^n$, $F_\text{us}^n$ and the corresponding features  $F_\text{d}^n$, $F_\text{s}^n$, as follows,
\begin{equation}
\label{eq:loss}
\mathcal{L}_\text{u}=\lambda_1||F_\text{ud}^n-F_\text{d}^n||^2+\lambda_2||F_\text{us}^n-F_\text{s}^n||^2,
\end{equation}
where $\lambda_1$ and $\lambda_2$ are weights that balance the depth estimation and scene  parsing. With this loss function, parameter updates take place within  block $n$ and the connecting coding branches, during which process  blocks 1 to $n-1$ remain unchanged.

The loss function of Eq.~(\ref{eq:loss}) is intuitively straightforward. However, using this loss to train each block in TargetNet turns out to be  time- and effort-consuming. This is because for training each block, we have to adjust the weights $\lambda_1$, $\lambda_2$ and the termination conditions, as the magnitudes of the feature maps and convergence rates vary block by block.

We therefore turn to another alternative that learns the features intertwined with the teachers, in aim to alleviate the cumbersome tuning process. For amalgamation in block $n$, we first obtain $F_\text{us}^n$ by passing the amalgamated features $F_\text{u}^n$ through the S-channel coding. We then \emph{replace} the features $F_\text{s}^n$ at the $n$-th block of SegNet with $F_\text{us}^n$, and obtain the consequent prediction $\hat{S}$ from the SegNet with $F_\text{us}^n$ being its features. This process is repeated in the same way to get the predicted depth map $\hat{D}$ from the DepthNet with $F_\text{ud}^n$ being its features.

In this way, we are able to write the loss as a function comprising only the final predictions $\hat{D}$, $\hat{S}$ and the predictions by original teachers ${D}$,${S}$, as follows,
\begin{equation}
\label{eq:lossFun}
\mathcal{L}_\text{u}=\lambda_1\mathcal{L}_\text{depth}(D,\hat{D})+\lambda_2\mathcal{L}_\text{seg}(S, \hat{S}),
\end{equation}
where the $\lambda_1$, $\lambda_2$ are fixed for all the blocks in TargetNet during training, and $\mathcal{L}_\text{depth}(\cdot)$  and $\mathcal{L}_\text{seg}(\cdot)$ are respectively those defined in  Eqs.~(\ref{eq:depthloss})~and~(\ref{eq:segloss}).



\subsubsection{Branch Out}
\label{sec:branchout}
As scene parsing and depth estimation are two closely related tasks, it is non-trivial to decide where to branch out the TargetNet into separate task-specific streams to achieve the optimal performances on both tasks simultaneously. Unlike conventional multi-branch models that choose to branch out at the boarder of encoder and decoder, we explore another option, which we find more effective.
After training the $N$ blocks of TargetNet using the loss function of Eq.~(\ref{eq:lossFun}), we also acquire the final losses for each block
$\{\mathcal{L}_\text{seg}^1, \mathcal{L}_\text{seg}^2,...,\mathcal{L}_\text{seg}^N\}$ and $\{\mathcal{L}_\text{depth}^1, \mathcal{L}_\text{depth}^2,...,\mathcal{L}_\text{depth}^N\}$. The blocks for branching out, $p_\text{seg}$ and $p_\text{depth}$, are taken to be:
\begin{equation}
\begin{split}
p_\text{seg}&=\arg\min_n{\mathcal{L}_\text{seg}^n}\\
p_\text{depth}&=\arg\min_n{\mathcal{L}_\text{depth}^n}
\end{split}
\end{equation}
where we set $N/2<n\le N$ to allow the branching out takes place only within the decoder.

Once the branch-out blocks $p_\text{seg}$ and $p_\text{depth}$ are decided, we remove those initial decoder blocks succeeding the last branch-out block. In the example shown in Fig.~\ref{fig:multi}, scene parsing  branches out later than depth estimation, and thus all the initial blocks after $p_\text{seg}$ are removed.
We then take the corresponding decoder blocks of the teachers as the branched-out blocks for TargetNet, as depicted by the upper- and lower-green stream in Fig.~~\ref{fig:multi}. In this way, we derive the final TargetNet architecture of one encoder and two decoders sharing some blocks, and fine-tune this model.



\begin{figure}
\centering
\includegraphics[scale = 0.55]{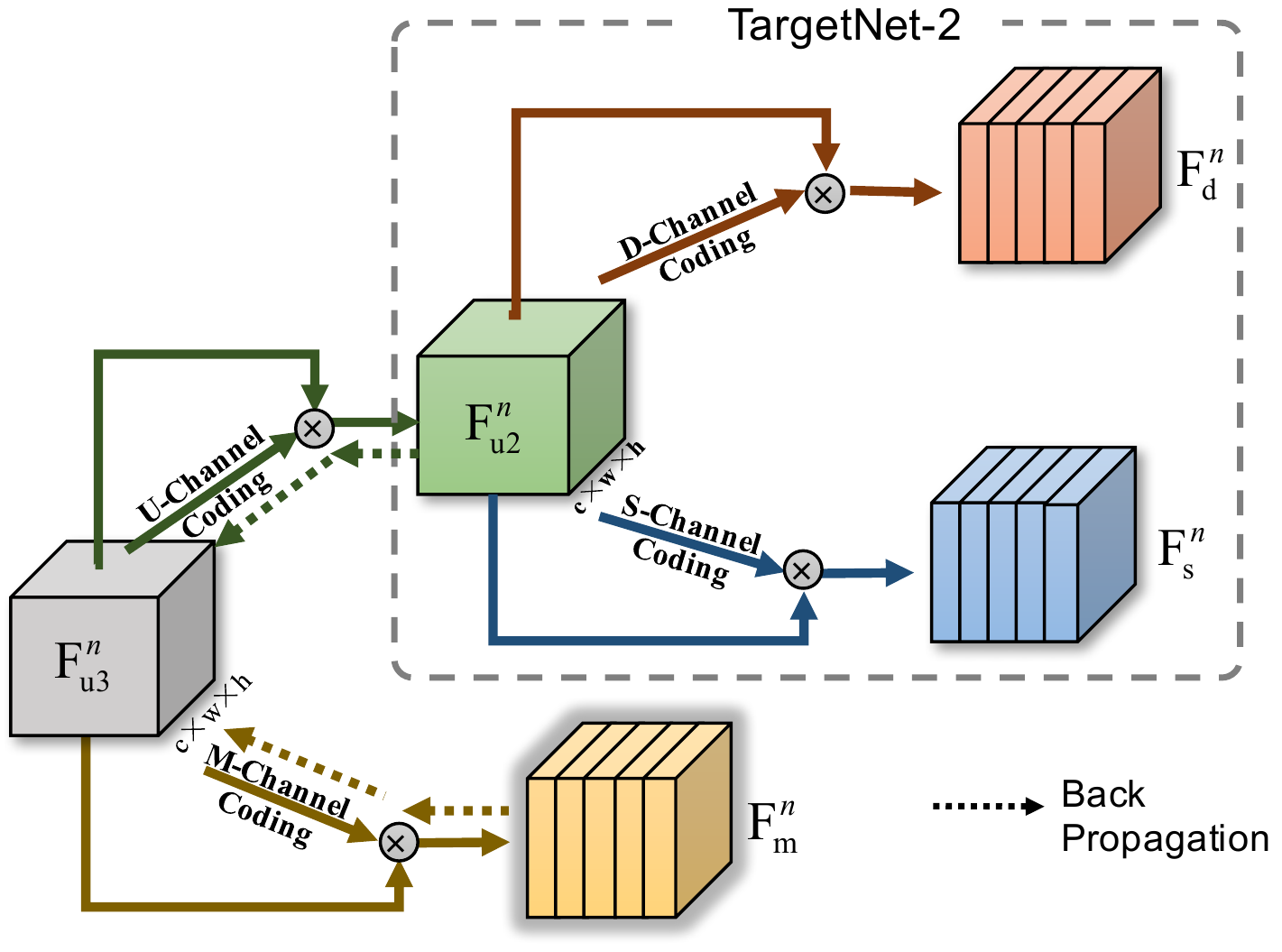}
\caption{ Knowledge amalgamation with TargetNet-2 and NormNet. The part in dash frame is from TargetNet-2 and keeps unchanged during training TargetNet-3.}
\label{fig:decoder3}
\vspace{-1em}
\end{figure}

\subsection{Learning from More Teachers}
\label{sec:moreTeachers}
In fact, the proposed training scheme is  flexible and not limited to amalgamating knowledge from two teachers only. We show here two approaches, an offline approach and an online one, to learn from three teachers. {We take surface-normal prediction, another widely studied pixel-prediction task, as an example to illustrate the three-task amalgamation. The proposed two approaches can be directly applied to amalgamation from an arbitrary number of teachers.}



The offline approach for three-task amalgamation is quite straightforward.
The block-wise learning strategy described in Sec.~\ref{sec:blockTrain} can be readily generalized here. Now instead of two coding channels we have three, wherein the third one \emph{M-channel coding} transfers the amalgamated knowledge to the normal-prediction task. The loss function is then taken to be
\begin{equation}
\begin{split}
\mathcal{L}_\text{u3}=\lambda_1\mathcal{L}_\text{depth}(D,\hat{D})+\lambda_2\mathcal{L}_\text{seg}(S, \hat{S}), +\lambda_3\mathcal{L}_\text{norm}(M,\hat{M}),
\end{split}
\end{equation}
where $\lambda_1$, $\lambda_2$ and $\lambda_3$ are the balancing weights.

The online approach works in an incremental manner. Assume we have already trained a TargetNet for parsing and depth estimation, which we call TargetNet-2 now for clarity. We would like the student to amalgamate the normal-prediction knowledge as well, and we call this three-task student  TargetNet-3. The core idea of this online approach is to treat TargetNet-2 itself as a pre-trained teacher, and then conduct amalgamation of TargetNet-2 and NormNet in the same way as done for the two-teacher amalgamation described in Sec.~\ref{sec:blockTrain}. Let $F_\text{u2}^n$ denote the features at block~$n$ of TargetNet-2, and let  $\{D_\text{u},S_\text{u}\}$ denote its predictions of depths and segmentation. Also, let $F_\text{u3}^n$ denote the features to be learned at the $n$-th block of TargetNet-3.

\begin{figure*}
\centering
\includegraphics[scale = 0.96]{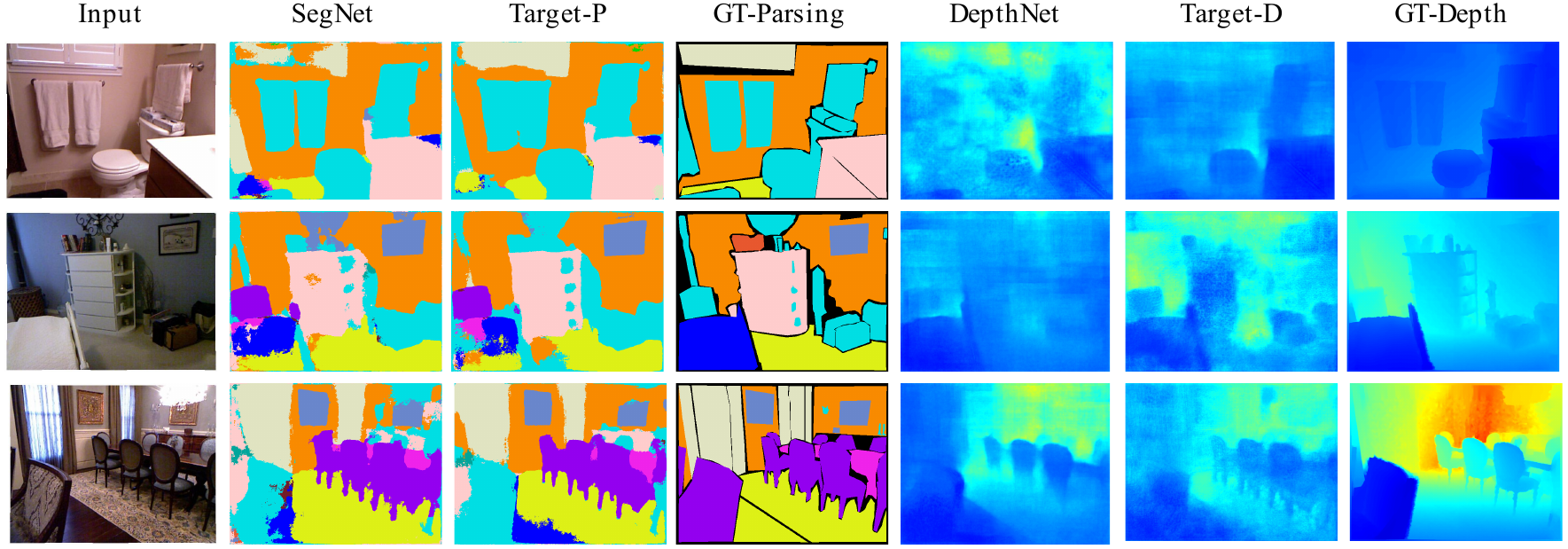}
\vspace{-0.0em}
\caption{Qualitative results on scene parsing and depth estimation on NYUDv2.
For both tasks, we compare the results of the teachers with those of the student, with Target-P and Target-D
denoting the results of the student on scene parsing and depth estimation respectively.}
\label{fig:exp}
\vspace{-1em}
\end{figure*}

We connect $F_\text{u3}^n$ and $F_\text{u2}^n$ via the \emph{U-channel coding} as depicted in Fig.~\ref{fig:decoder3}.
On the normal-prediction side, we pass $F_\text{u3}^n$  through the M-channel coding to produce the projected features $F_\text{m}^n$,  replace the corresponding features of NormNet, and obtain the norm prediction $\hat{M}$. On the TargetNet-2 side, we project  $F_\text{u3}^n$ through U-channel coding to obtain $F_\text{u2}^n$, repeat the same process and obtain predictions  $\{\hat{S}_\text{u},\hat{D}_\text{u}\}$. The  loss function is then taken to be
\begin{equation}
\mathcal{L}_\text{u3}=\lambda_1\mathcal{L}_\text{norm}(\hat{M},M)+\lambda_2\mathcal{L}_\text{u2},
\label{eq:onlineMulti}
\end{equation}
where $\lambda_1$ and $\lambda_2$ are the weights. As shown in Fig.~\ref{fig:decoder3}, we freeze TargetNet-2 during this process since it is treated as a teacher. Thus, Eq.~(\ref{eq:onlineMulti}) will update the parameters of U-channel coding but not those within TargetNet-2.


%

\section{Experiments and Results}
Here we provide our experimental settings and results. More results can be found in our supplementary material.

\subsection{Experimental Settings}
\textbf{Datasets.}
The NYUDv2 dataset~\cite{Silberman:ECCV12}
provides 1,449 labeled indoor-scene RGB images with both parsing annotations and Kinect depths,
and 407,024 unlabeled ones, of which 10,000  are used to train TargetNet.
Besides the scene parsing and depth estimation task, we also train TargetNet-3 that conducts
surface normal prediction on this dataset.
The surface normal ground truths are computed from the depth maps
using the neighboring pixels' cross products.


Cityscapes~\cite{Cordts2016Cityscapes} is a large-scale dataset for semantic urban scene understanding.
It provides RGB images and stereo disparity ones,
collected over 50 different cities spanning several months, with overall 19 semantic classes being annotated.
The finely annotated part consists of 2,975 images for training, 500 for validation, and 1,525 for test.
Disparity images provided by Cityscape are used as depth maps, with bilinear interpolation filling the holes.

\begin{table*}
\centering
\scriptsize
\caption{Comparative results of the teacher networks and the student  with different branch-out blocks
on the NYUDv2 dataset. {Decoder\_b$n$} denotes the TargetNet that branches out at the block $n$ of the decoder, and TeacherNet contains SegNet and DepthNet. The final student network branches out the depth estimation at block~4, denoted by Target-D,
and scene parsing at block~5 denoted by Target-P.}
\vspace{1mm}
\label{tab:nyutarget2}
\begin{tabular}{l|p{12mm}<{\centering}|p{13mm}<{\centering}p{13mm}<{\centering}|p{10mm}<{\centering}p{10mm}<{\centering}|p{13mm}<{\centering}p{13mm}<{\centering}p{13mm}<{\centering}}
\hline
\textbf{Method }  & \textbf{Params} & \textbf{mean IoU}   & \textbf{Pixel Acc.}& \textbf{abs rel} & \textbf{sqr rel}&\bm{$\delta<1.25$} &\bm{$\delta<1.25^2$}&\bm{$\delta<1.25^3$}   \\ \hline
TeacherNet& $\sim$55.6M& 0.448 & 0.684 & 0.287   & 0.339   & 0.569 & 0.845 & 0.948 \\
\hline
Decoder\_b1 &$\sim$36.9M&0.447 & 0.684 & 0.276   & 0.312   & 0.383 & 0.753 & 0.939\\
Decoder\_b2 &$\sim$31.0M&0.451 & 0.684 & 0.259   & 0.275   & 0.448 & 0.799 & 0.952\\
Decoder\_b3 &$\sim$28.3M&0.451 & 0.684 & 0.260   & 0.277   & 0.448 & 0.796 & 0.951\\
\bf{Decoder\_b4} (Target-D) &$\sim$28.0M&0.452 & 0.683 & \bf{0.252}   & \bf{0.257}  & \bf{0.544} & \bf{0.847} & \bf{0.959}\\
\bf{Decoder\_b5} (Target-P) &$\sim$27.8M&\bf{0.458} &\bf{0.687} & 0.256   & 0.266   & 0.459 & 0.810 & 0.956\\ \hline
\end{tabular}
\vspace{-1em}
\end{table*}

\textbf{Evaluation Metrics.}
To qualitatively evaluate the depth-estimation performance, we use the standard metrics as done in~\cite{eigen2014depth}, including absolute relative difference (abs rel), squared relative difference (sqr rel) and the percentage of relative errors inside a certain threshold $t$.

For scene parsing, we use both the pixel-wise accuracy (Pixel Acc.) and the mean Intersection over Union (mIoU).

For surface normal prediction, performance is measured by the same matrix as~\cite{eigen2015predicting}: the mean and median angle from the groundtruth across all unmasked pixels, as well as the percent of vectors whose angles fall within three thresholds.

\textbf{Implementation Details.}
The proposed method is implemented using TensorFlow with a NVIDIA M6000 with 24GB memory. We use the poly learning rate policy as done in~\cite{liu2015parsenet:}, and set the base learning rate to 0.005, the power to 0.9, and the weight decay to $4e-6$. We perform data augmentation with random-size cropping to $640\times 400$ for all datasets. Due to limited physical memory on GPU cards and to assure effectiveness, we set the batch size to be 8 during training. The encoder and decoder of TargetNet are initialized with parameters pre-trained on ImageNet.
In total 2 epochs are used for training each block of the Target on both NYUDv2 and Cityscape.

\subsection{Experimental Results}
\textbf{Performance on NYUDv2.}
We show in Tab.~\ref{tab:nyutarget2} the qualitative results of the TeacherNet (SegNet and DepthNet) and those of the student network with different branch-out blocks, denoted by {Decoder\_b1-5}, on the testset of NYUDv2. For example, Decoder\_b1 corresponds to the student network branched out at decoder block~1.
Our final TargetNet branches out depth estimation at block~4 and scene parsing at block~5, which give us the best performances for the two tasks.
As can be seen, the student networks consistently outperform the teachers in terms of all evaluation measures, which validates the effectiveness of the proposed method. As for the network scale, the more blocks of the features  are amalgamated, the smaller TargetNet is, which means that branching out at a later stage of the decoder leads to smaller networks, as shown by the number of parameters of Decoder\_b1-5.
The final student network, which  branches out at blocks~4~and~5 respectively for depth and scene, yields a size that is about half of the teachers.

The qualitative results are displayed in Fig.~\ref{fig:exp}, where the results of student,  {Target-P} and {Target-D},
are more visually plausible than those of the teacher networks.




\begin{figure}
\centering
\includegraphics[scale = 0.72]{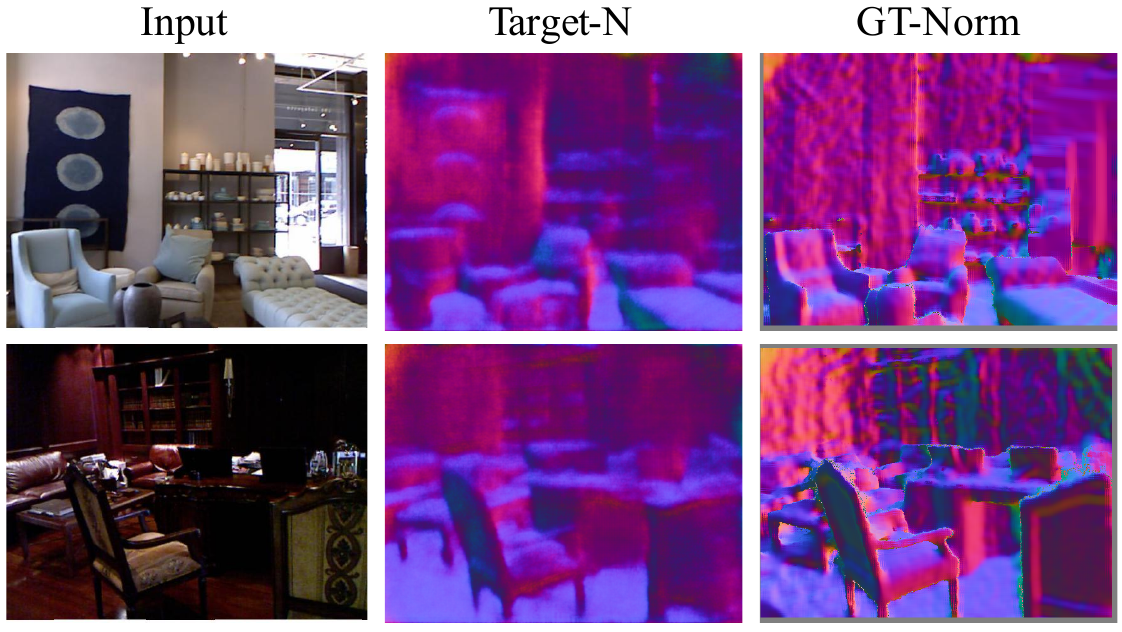}
\caption{Qualitative results on surface normal prediction on NYUDv2.
The {Target-N} column corresponds to the results of the proposed TargetNet-3.}
\label{fig:norm}
\end{figure}

\begin{table}[]
\scriptsize
\centering
\caption{Comparative results of the TeacherNet (SegNet, DepthNet and NormNet), Target-2 and Target-3 on NYUDv2.}
\vspace{1mm}
\label{tab:target-3}
\begin{tabular}{p{11.8mm}|p{4.8mm}<{\centering}|p{3mm}<{\centering}p{4mm}<{\centering}|p{3mm}<{\centering}p{6mm}<{\centering}|p{4.6mm}<{\centering}p{3.7mm}<{\centering}p{4mm}<{\centering}}
\hline
       &                       & \multicolumn{2}{c|}{\bf{rel diff}} & \multicolumn{2}{c|}{\bf{Angle Dist.}}          & \multicolumn{3}{c}{\bf{With} \bm{$t^\circ$} \bf{Deg.}}                                           \\
\bf{Method} & \bf{mIOU} & \bf{abs} & \bf{sqr}  & \bf{Mean} & \bf{Median}                &   $11.25^\circ$   & $22.5^\circ$&  $30^\circ$      \\ \hline
TeacherNet       &  0.448   &   0.287  & 0.339 & 37.88  &  36.96 &  0.236 &\bf{0.450} &  0.567  \\
TargetNet-2      &  0.458   &   0.252  & 0.257 & $-$    &  $-$   &  $-$   &  $-$    &  $-$    \\
TargetNet-3 &  \bf{0.459} &\bf{0.243} & \bf{0.255}& \bf{35.45} & \bf{34.88}&\bf{0.237} & 0.448& \bf{0.585}\\ \hline
\end{tabular}
\end{table}

\begin{table}[]
\centering
\scriptsize
\caption{Comparative results of TargetNet and state-of-the-art methods on scene parsing and depth estimation on NYUDv2.}
\vspace{1mm}
\label{tab:compare}
\begin{tabular}{l|c|cc|cc}
\hline
\textbf{Method} &\bf{Params} &\textbf{mIOU} & \textbf{PA}& \bf{abs rel} &\bf{sqr rel}\\ \hline

Xu et al.~\cite{xu2017multi-scale}    &$-$& $-$   & $-$ & 0.121&$-$  \\
FCN-VGG16~\cite{long2015fully}   &$\sim$134M& 0.292  & 0.600&$-$&$-$\\
RefineNet-L152~\cite{Ne2018lwrefinenet:}&$\sim$62M & 0.444& $-$&$-$&$-$\\
RefineNet-101~\cite{lin2017refinenet:}    &$\sim${118}M& 0.447   & 0.686&$-$&$-$ \\ \hline
Gupta et al.~\cite{gupta2014learning} &$-$& 0.286& 0.603&$-$&$-$ \\
Arsalan et al.~\cite{mousavian2016joint} &$-$& 0.392& 0.686&0.200&0.301 \\
PADNet-Alexnet~\cite{xu2018pad-net:} &$>$50M& 0.331   & 0.647& 0.214 & $-$\\
PADNet-ResNet~\cite{xu2018pad-net:} &$>$80M& 0.502   & 0.752 & 0.120 & $-$ \\
\hline
TargetNet & $\sim$28M  &0.459& 0.688 & 0.124&0.203 \\
\hline
\end{tabular}
\end{table}

For surface normal prediction, we train the TargetNet-3 described in Sec.~\ref{sec:moreTeachers}
in the offline manner, and show the results in Tab.~\ref{tab:target-3}. With the surface normal information amalgamated in TargetNet-3, the accuracies of scene parsing and depth estimation increase even further and exceed those of the TargetNet-2. In particular, compared to the enhancement of scene parsing, that of  depth estimation is more significant due to its tight relationship with surface normal. The visual results of the surface normal prediction are depicted in Fig.~\ref{fig:norm}.

We also compare the performance of TargetNet with those of the state-of-the-art models. The results are shown in Tab.~\ref{tab:compare}. TargetNet is trained with three tasks. With the smallest size and no access to human-labelled annotations,  TargetNet yields results better than all but one compared methods, PADNet-ResNet~\cite{xu2018pad-net:}, for which the size is almost three times as that of TargetNet.


\textbf{Performance on Cityscape.}
We also conduct the proposed knowledge amalgamation on the Cityscape dataset, and show the quantitative results in Tab.~\ref{tab:city}.
We carry out the experiments in the same way as done for NYUDv2. Results on this outdoor dataset further validate the effectiveness of our approach, where
the amalgamated network again both surpasses the teachers on their own domains.

\begin{table}[]
\scriptsize
\centering
\caption{Comparative results of the TeacherNet (SegNet and DepthNet) and the student on the Cityscape dataset. {Target-D} and {Target-P} denote the results of TargetNet for parsing and depth estimation respectively.}
\vspace{1mm}
\label{tab:city}
\begin{tabular}{p{15mm}|cc|p{12mm}<{\centering} p{12mm}<{\centering}}
\hline
\textbf{Method} & \textbf{mean IOU} & \textbf{Pixel Acc.}& \textbf{abs rel}& \textbf{sqr rel} \\ \hline
TeacherNet    & 0.521   & 0.875 & 0.289& 5.803\\
Target-P    & \bf{0.535}   & \bf{0.882} & 0.240 & 3.872\\
Target-D    & 0.510   & 0.882 & \bf{0.224} & \bf{3.509}\\

\hline
\end{tabular}
\vspace{-1mm}
\end{table}
\section{Conclusion}
In this paper, we investigate a novel knowledge amalgamation task, which aims to learn a versatile student model from pre-trained teachers specializing in different application domains, without human-labelled annotations. We start from  a pair of teachers, one on scene parsing and the other on depth estimation, and propose an innovative strategy to learn the parameters of the student intertwined with the teacher. We then present two options to generalize the training strategy for more than two teachers. Experimental results on several benchmarks demonstrate that student model, once learned, outperforms the teachers in their own expertise domains. In our future work, we will explore knowledge amalgamation from teachers with different architectures, in which case the main challenge is to bridge the semantic gaps between the feature maps of the  teachers.

\section*{Acknowledgement}
This work is supported by  National Key Research and Development Program (2016YFB1200203), National Natural Science Foundation of China (61572428,U1509206), Fundamental Research Funds for the Central Universities (2017FZA5014),  Key Research and Development Program of Zhejiang Province (2018C01004), and the Startup Funding of Stevens Institute of Technology.


\end{document}